# Multi-Amateur Contrastive Decoding for Text Generation

Jaydip Sen
*Department of Data Science*
*Praxis Business School*
Kolkata, INDIA
email: jaydip.sen@acm.org

Subhasis Dasgupta
*Department of Data Science*
*Praxis Business School*
Kolkata, INDIA
email: subhasisdasgupta1@acm.org

Hetvi Waghela
*Department of Data Science*
*Praxis Business School*
Kolkata, INDIA
email: waghelah@acm.org

*Abstract*— Contrastive Decoding (CD) has emerged as an effective inference-time strategy for enhancing open-ended text generation by exploiting the divergence in output probabilities between a large expert language model and a smaller amateur model. Although CD improves coherence and fluency, its dependence on a single amateur restricts its capacity to capture the diverse and multifaceted failure modes of language generation, such as repetition, hallucination, and stylistic drift. This paper proposes Multi-Amateur Contrastive Decoding (MACD), a generalization of the CD framework that employs an ensemble of amateur models to more comprehensively characterize undesirable generation patterns. MACD integrates contrastive signals through both averaging and consensus penalization mechanisms and extends the plausibility constraint to operate effectively in the multi-amateur setting. Furthermore, the framework enables controllable generation by incorporating amateurs with targeted stylistic or content biases. Experimental results across multiple domains, such as news, encyclopedic, and narrative, demonstrate that MACD consistently surpasses conventional decoding methods and the original CD approach in terms of fluency, coherence, diversity, and adaptability, all without requiring additional training or fine-tuning.

*Keywords—Contrastive Decoding, Open-ended Text Generation, Language Models, Decoding Strategies, Controllable Generation, Model Ensembles, Inference-time Optimization, Natural Language Generation.*

## I. Introduction

Open-ended text generation has become one of the most impactful uses of modern language models, powering applications in areas like creative writing, chatbots, summarization, question-answering, and personalized content delivery. Autoregressive language models, particularly those built on the transformer architecture and trained on massive datasets, have shown an impressive ability to generate fluent, contextually relevant text. Models such as GPT [1], OPT [2], and T5 [3] now produce language that, in some scenarios, rivals or even surpasses human writing. Still, the effectiveness of these models hinges not just on their architecture and pretraining, but also on how their outputs are decoded during inference. The decoding strategy, essentially, how the model's predictions are turned into a final sequence, plays a crucial role in shaping the coherence, relevance, and quality of the generated text.

Decoding strategies in language generation generally fall into two main types: deterministic and probabilistic. Deterministic methods, such as greedy decoding and beam search, prioritize selecting the highest probability tokens at each step. While they often produce grammatically sound and consistent outputs, the results can feel overly rigid, repetitive, or lacking in variation. On the other hand, probabilistic approaches like top-$k$ sampling [4], nucleus sampling [5], and typical decoding [6] introduce randomness by drawing from a subset of the model's predictions. These techniques promote diversity and creativity in the output, but often at the cost of coherence or topical consistency, issues that become more pronounced in longer texts. This creates a persistent challenge: balancing diversity and relevance, as pushing too far in either direction can result in content that strays from the original prompt or becomes disjointed.

To overcome these limitations, a recent and promising approach involves the use of *contrastive decoding* (CD), which reconceptualizes decoding as an optimization task [7]. Instead of solely maximizing the probability under a single language model, the CD decoding evaluates the difference in likelihoods between two language models of different capacities: a large model, the *expert*, and a smaller model, the *amateur*. The intuition behind this approach is that undesirable generation patterns, such as repetition, redundancy, and incoherence, are more prevalent in weaker models. By suppressing tokens favored by the amateur and emphasizing those favored by the expert, the CD framework achieves a fine balance between coherence, fluency, and informativeness. Importantly, this approach does not require any additional training, fine-tuning, or architectural modification. It operates entirely with frozen, pre-trained models and can be applied across domains and model families.

The core CD objective is formulated as the log-likelihood difference between the expert and amateur models, with the generated token constrained to be plausible under the expert. To implement this constraint, a filtering mechanism is employed to exclude tokens whose expert-assigned probability falls below a specified threshold relative to the most probable token at each time step. This design ensures that generated sequences remain both plausible and semantically rich, while systematically filtering out repetitive or incoherent continuations.

Empirical studies show that CD can significantly enhance the quality of the generated text, particularly with respect to coherence, topic relevance, and user preference, when compared to more conventional decoding methods [7]. Techniques like top-$k$ sampling, nucleus sampling, greedy decoding, and beam search often face difficulties balancing fluency and diversity, leading to outputs that are either overly repetitive or semantically disjointed, especially in longer passages. CD addresses these shortcomings by leveraging the output of a weaker language model to filter out less promising token choices, effectively guiding the stronger model toward more meaningful continuations. This strategy has demonstrated consistent benefits across a wide range of text domains, including encyclopedic content, news reporting, and fiction. Notably, CD tends to perform better on metrics like MAUVE, which evaluates how closely the distribution of generated text aligns with human-written examples and also shows a marked reduction in *n*-gram repetition [8].

Despite its strengths, the CD framework has notable limitations. One of the main concerns is its reliance on a single weaker, or *amateur* model to guide the decoding process. While effective in some cases, this model may only detect a narrow slice of problematic behaviors in generated text. Issues like excessive repetition, syntactic breakdowns, factual inaccuracies, or awkward stylistic shifts often require more effective evaluation than a single model can provide. Assuming that one lightweight model can account for the full range of undesirable outputs may constrain the flexibility and robustness of the approach. This may, potentially limit its performance in more diverse or challenging generation scenarios.

Moreover, the original CD method lacks the flexibility required for specialized generation tasks such as domain-specific adaptation, style transfer, or bias mitigation. In practical applications, it is often essential to exert fine-grained control over the generated content, whether to preserve a formal tone, steer clear of inappropriate language, or adhere to the conventions of a particular genre. However, current decoding strategies, including CD, do not offer built-in mechanisms to enforce such constraints. They typically require external intervention or retraining to achieve these goals.

To address these gaps, the present work introduces an enhanced decoding framework termed *Multi-Amateur Contrastive Decoding* (MACD). This method generalizes the original CD formulation by incorporating multiple amateur models, each selected to reflect specific degenerative tendencies or failure modes. By aggregating the likelihoods from several amateur models, the decoding process can simultaneously penalize a broader array of undesirable characteristics. These models may include, small-scale LMs, *n*-gram-based approximations, distilled variants, or models trained on biased or informal data. The diversity of the amateur ensemble allows for a more comprehensive suppression of errors during decoding.

To integrate the amateur ensemble effectively, two complementary aggregation strategies are introduced: (i) *mean penalization*, which computes the contrastive loss by averaging across the outputs of all amateur models, and (ii) *consensus penalization*, which applies a penalty only to those tokens consistently disfavored by the majority of amateurs. These approaches strike a balance between robustness and computational efficiency, offering adaptable means of suppressing undesirable generation patterns. Moreover, the plausibility constraint is adapted to the ensemble setting by introducing joint thresholds that account for both the expert model's confidence and the collective judgment of the amateurs. This *ensemble-aware* constraint enhances the reliability of token selection while guarding against excessive pruning of valid continuations.

The MACD framework is also evaluated for its applicability in controlled generation. By curating amateur models that exemplify specific unwanted styles, such as informal tone, toxicity, or hallucination, it becomes possible to steer the generation process toward outputs that are cleaner, and more neutral. This enhancement allows MACD to serve not only as a general-purpose decoding strategy but also as a plug-in mechanism for zero-shot controllable generation without requiring additional model retraining or fine-tuning.

At a conceptual level, the MACD decoding objective shares strong ties with ideas like *pointwise mutual information* (PMI) and *pragmatic informativeness* [6]. It can be understood as a way of highlighting tokens that stand out to a stronger model when compared to simpler baselines. This contrast acts as a signal of what the expert model finds meaningful or distinctive. The approach also echoes well-known principles in human communication, such as those proposed by Li et al., which suggest that effective messages should be both relevant and not repetitive [7]. From this perspective, MACD encourages generated text to be not just fluent and plausible, but also genuinely informative.

The remainder of this paper is organized as follows. Section II reviews related work in CD, ensemble-based inference methods, and decoding strategies for controlled and diverse text generation. Section III provides background on the original CD framework, detailing its theoretical foundations and algorithmic formulation. Section IV introduces the proposed MACD approach, outlines the ensemble design and consensus-based scoring strategies, and presents the extension of plausibility constraints in the multi-amateur setting. Section V highlights the experimental results, including automatic and human evaluations across diverse domains, and compares them with strong baselines. Section VI concludes the paper by summarizing the key findings, discussing implications for controllable generation, and identifying future research directions.

## II. RELATED WORK

Deterministic approaches often stick too closely to the most likely output, resulting in repetitive or rigid language. On the other hand, sampling-based techniques allow for more variety but sometimes lose coherence or drift from the original prompt. These limitations have led to a wave of research into more flexible decoding strategies, ones that can respond to context, filter poor predictions, and make smarter use of the model's knowledge. The following discussion looks at some of the most influential recent developments in this area, identifying their contributions as well as the gaps they leave behind. These gaps help motivate the need for a new approach, which is where the MACD framework comes in.

Su & Collier challenged the assumption that token representations in language models are inherently anisotropic and introduced *contrastive search* as a way to improve the quality of generated text [9]. Their approach reached near-human performance across several languages, all without requiring further model training. However, the proposed scheme by the authors did not directly address concerns about controlling the generation behavior or reducing specific types of failure, such as incoherence or factual errors.

Garces Arias et al. proposed an extension that added an adaptive degeneration penalty, guided by model uncertainty, to boost both creativity and diversity in the generated text [10]. While this refinement led to improvements across various models and datasets, it also introduced greater complexity due to the need for accurate uncertainty estimation.

An et al. tackled the issue of *exposure bias* in neural text generation by introducing a contrastive learning framework aimed at improving sequence-level training [11]. Their method achieved impressive results, setting new benchmarks in the field. However, there's a catch: the approach requires modifications to the training process, which can be a bit of a hurdle for those looking to work with pre-trained models without starting from scratch.

Yona et al. proposed *contrastive input decoding* (CID) as a tool for uncovering how language models respond differently to slight changes in input. By generating text from opposing prompts, CID helps detect context-specific biases that might otherwise go unnoticed [12]. While it's a valuable method for analyzing model behavior, its role is mostly diagnostic. It doesn't offer a direct path to improving the quality or reliability of the generated text.

Wiher et al. conducted a detailed evaluation of several decoding strategies across a variety of natural language generation tasks [13]. Their analysis showed that the effectiveness of any given method is highly dependent on the context, it's not a one-size-fits-all situation. While their work highlighted the importance of developing more flexible and adaptable decoding techniques, it didn't go as far as offering new algorithms to meet those needs.

Jimale et al. introduced *DecoStrat*, a modular framework integrating multiple decoding strategies, including *minimum Bayes-risk* (MBR) decoding, to enhance data-to-text generation [14]. It demonstrated its effectiveness on the MultiWOZ dataset [15]. However, its focus on structured data limits its applicability to open-ended text generation tasks.

Yuan et al. introduced a method that blends speculative decoding with contrastive decoding by using a smaller model to suggest possible next tokens, which are then confirmed by a larger and more powerful model [16]. The idea is to boost both the speed and quality of text generation. While the approach shows promising gains in efficiency without sacrificing output quality, it depends heavily on how well the smaller and larger models align, something that isn't always guaranteed.

Liu et al. proposed *knowledge-infused decoding* (KID), a technique that integrates external knowledge into the decoding process by employing a dynamic memory structure to guide token selection [17]. The primary objective of the scheme is reducing hallucinations and enhancing factual accuracy. The method incorporates reinforcement learning to adaptively modulate the influence of the knowledge memory during generation. Although KID demonstrates improvements in factual consistency, its performance is contingent on the availability of structured external knowledge sources. Moreover, it incurs additional computational overhead.

Zeng et al. proposed *CoDec*, a cooperative decoding framework that brings together the strengths of *neural machine translation* (NMT) systems and *large language models* (LLMs) [18]. In this setup, the NMT system produces an initial draft, which is then refined by the LLM to improve fluency and naturalness. By combining the accuracy of traditional NMT with the creative flexibility of LLMs, CoDec manages to boost translation quality. However, because it's tailored specifically for translation tasks, its usefulness in more general text-generation contexts remains limited.

Wang et al. proposed *semi-autoregressive translation* (SAT), which improves decoding efficiency by allowing the parallel generation of token groups rather than strictly sequential generation [19]. While it offers significant speedups over traditional autoregressive models with only a slight drop in quality, its fixed grouping strategy may fail to adapt dynamically to linguistic complexity. Moreover, its utility is largely demonstrated in translation tasks, and its generalization ability in open-ended generation remains unexplored.

Li and Jurafsky explored an innovative way to make translations more accurate and meaningful. Instead of just picking the most likely sentence, their method looks for translations that share the most information with the original, essentially trying to stay true to both the content and the intent [20]. They do this by generating a list of possible outputs and then using a second model, which works in the reverse, to figure out which ones make the most sense. It's a smart approach that leads to richer, more relevant translations. However, it is computationally intensive due to its reliance on the execution of an extra model just to re-rank the options.

Vijayakumar et al. introduced *diverse beam search* (DBS), a decoding strategy that enhances output diversity by incorporating diversity constraints within groups of candidate sequences during beam search [21]. This method effectively mitigates the issue of generating near-identical sequences, particularly in visually grounded tasks like image captioning and visual question generation. However, while DBS improves diversity, it may compromise output quality if diversity constraints are over-emphasized.

Sen et al. propose several novel decoding schemes that advance the efficiency, robustness, and quality of LLM inference [22-25]. The *contextual contrastive search* (CCS) introduces a decoding mechanism that dynamically modulates temperature and re-ranking based on context and entropy, enhancing coherence in long-form generation [22]. The *adaptive semantic-aware typicality sampling* (ASTS) augments typicality sampling with entropy-tuned thresholds and semantic alignment filters to improve fluency and informativeness without increasing computational burden [23]. The *confidence-modulated speculative decoding* (CM-ASD) adaptively adjusts drafting lengths and verification criteria using entropy-based confidence signals, significantly accelerating decoding while maintaining quality across uncertain inputs [24]. Building on these, the *hierarchical verification tree* (HVT) framework restructures speculative beam decoding into a priority-driven tree, enabling early pruning of low-probability candidates and achieving superior performance in term of latency, energy efficiency, and output fidelity [25].

*Comparison with Recent Decoding Frameworks*: Recent advancements in decoding techniques have attempted to balance the generation quality, controllability, and computational efficiency. For instance, speculative decoding [16] improves the inference speed by leveraging a small draft model to propose token, which are then validated by a larger expert model. However, it has limited controllability, and its performance depends on the alignment between the draft and verifier models. In contrast, the proposed MACD framework introduces ensemble-based penalization using multiple amateur models enabling targeted suppression of undesirable generation patterns. Moreover, the MACD supports domain- and bias-specific filtering in zero-shot, inference-only setting, without the need for retraining or architectural changes. Energy-based decoding frameworks adopt scoring functions over sequences that reflect both generation likelihood and external constraints [26]. While these schemes produce high-quality text, they often suffer from high computational costs. MACD, on the other hand, optimizes the computational cost by leveraging precomputed token-level probabilities and supports parallelized inference across the amateur ensemble. Prompt editing and constrained generation methods [27] allow flexible steering of outputs via attribute models or gradient-based perturbations. However, these often involve backpropagation during inference, which causes significant latency. MACD avoids this by using forward-pass-only amateur models to shape generation strategies, which ensures stability and compatibility with expert models. Reinforcement learning-based decoding has also been explored by several researchers to optimize for task-specific objectives, such as factual accuracy or style [28]. However, these methods usually require task-aligned reward functions and extensive training. MACD operates in a fully zero-shot, inference-time paradigm, and supports plug-and-play control, making it more generalizable across domains and tasks. Overall, MACD strikes a balance between controllability, generality, and computational cost, and offers a lightweight, ensemble-driven decoding strategy that requires minimal fine-tuning and inference-time optimization.

### III. BACKGROUND: CONTRASTIVE DECODING

This section introduces the key concepts behind the development of more effective decoding strategies in open-ended text generation. It focuses on the CD framework, which improves the generation quality by comparing the prediction behaviors of two autoregressive language models with different capacities. The CD is based on the idea that smaller models are more likely to generate low-quality or degenerate outputs with excessive repetition, syntactic issues, or semantic drift. By contrasting the outputs of a large expert model with those of a smaller amateur model, CD creates a scoring mechanism that favors more fluent, coherent, and contextually appropriate tokens.

This section provides a detailed explanation of the mathematical formulation and decoding algorithm of the CD framework and compares its performance to traditional methods like greedy decoding, beam search, and sampling. It also discusses the limitations of CD, particularly its reliance on a single amateur model, which motivates the more generalized multi-amateur approach presented in Section IV.

#### A. Contrastive Decoding Framework

The CD is a recently proposed inference-time algorithm designed to enhance the quality of open-ended text generation by leveraging the behavioral asymmetry between two *autoregressive language models* (ALMs) of differing capacities: a high-capacity *expert* model and a smaller, less expressive *amateur* model. Unlike traditional decoding methods, which rely solely on a single model's probability distribution, CD introduces a *contrastive scoring mechanism* to prioritize fluent, coherent, and contextually appropriate tokens while suppressing those that are typically over-represented in lower-quality generations.

At the core of this method lies the autoregressive formulation of language modeling. Given an input sequence $x = (x_1, x_2, ..., x_T)$, an autoregressive LM estimates the probability of the sequence as presented in (1).

$$P(x) = \prod_{t=1}^{T} P(x_t|x_{<t}) \qquad (1)$$

In (1), each token $x_t$ is predicted conditioned on its preceding context $x_{i<t}$. During the generation, the model produces one token at a time, iteratively updating the context until a stopping condition is met. While this framework supports coherent local structure, the quality of full-sequence outputs heavily depends on the decoding strategy.

*CD* modifies the standard decoding process by introducing a *contrastive score* that ranks candidate tokens based on their differential likelihood under the expert and amateur models. For a candidate token $x_t$, the contrastive score is computed using (2).

$$S(x_t) = logP_E(x_t|x_{<t}) - \alpha * logP_A(x_t|x_{<t}) \qquad (2)$$

In (2), $P_E$ and $P_A$ denote the expert and amateur conditional probabilities, and $\alpha \geq 0$ is a *contrastive coefficient* that regulates the influence of the amateur model. Tokens that are preferred by the expert but disfavored by the amateur receive higher scores and are thus more likely to be selected during decoding.

To ensure that the decoding process remains anchored to linguistically plausible outputs, a plausibility constraint is introduced. The candidate token set $v_t \subset v$ ($v_t$ and $v$ are the candidate token set and the vocabulary set, respectively) is filtered by retaining only those tokens whose expert log-probability falls within a margin $\delta$ of the highest-scoring token as in (3).

$$logP_E(x_t|x_{<t}) \geq logP_E(x_t^*|x_{<t}) - \delta \qquad (3)$$

In (3), $x_t^* = arg\max_{x \in v} P_E(x_t|x_{<t})$, and $\delta \geq 0$ is a threshold controlling the strictness of the filter. This constraint excludes low-confidence expert predictions, thereby maintaining output quality.

At each decoding step $t$ the algorithm proceeds as follows:

1. The expert model computes conditional probabilities $P_E(x_t|x_{<t})$ over the vocabulary.
2. A plausibility filter is applied to retain only tokens whose log probabilities are within a threshold $\delta$ of the most likely token.
3. The amateur model computes probabilities $P_A(x_t|x_{<t})$ over the filtered set.
4. The contrastive score $S(x_t)$ is computed for each candidate token as: $logP_E(x_t|x_{<t}) - \alpha * logP_A(x_t|x_{<t})$
5. The token with the highest contrastive score is selected, and the sequence is extended.

The process repeats iteratively until a stopping condition, such as an end-of-sequence token or maximum length, is met.

#### B. Limitations of Contrastive Decoding

While CD represents a meaningful step forward in enhancing text generation quality, it still has its limitations. The method hinges on comparing the outputs of two models: the expert, which is more capable, and the amateur, which is less reliable. The contrastive approach does improve aspects like fluency, coherence, and relevance in the generated text. However, its reliance on just one single amateur model limits its capability. The single amateur model doesn't capture the full range of issues that can occur in text generation, such as repetitive phrases or lack of coherence due to biases or inconsistencies in the model. So, even though CD offers improvements over traditional methods, it still struggles with the complexity and variety of errors in text generation.

Moreover, the effectiveness of the contrastive score, shaped by the expert and amateur models, heavily depends on the choice of the amateur model. A single amateur model might not capture the full spectrum of undesirable generation behaviors, especially in more complex contexts. For example, while the expert model may generate high-quality predictions, it could still produce subtle yet significant flaws in the output, flaws that may not be sufficiently penalized by the amateur model. This highlights the limitations of relying on just one weaker model to detect and correct all potential problems.

These limitations motivate the need for a more generalized approach, one that incorporates multiple amateur models to better address the wide variety of generation pitfalls that may arise in open-ended text generation tasks. This gap serves as the foundation for the proposed MACD framework, which aims to enhance the robustness and adaptability of the decoding process by capturing a broader range of undesirable behaviors.

## IV. MULTI-AMATEUR CONTRASTIVE DECODING

The CD framework presented in Section III marked a significant improvement in generation quality by leveraging the asymmetry between a high-capacity expert model and a weaker amateur model. This contrastive approach effectively suppresses outputs symptomatic of low-quality generation such as excessive repetition, syntactic breakdowns, or semantic drift. However, the reliance on a single amateur model introduces a fundamental limitation: that constrains the decoding process to a narrow view of undesirable generation behavior. In practice, generation flaws are multifaceted and vary widely across tasks, domains, and input contexts. A single amateur model may fail to simultaneously detect or penalize disjoint issues such as factual hallucinations, grammatical inconsistencies, or stylistic deviations. Moreover, the choice of the amateur model is often arbitrary or general-purpose, which limits the adaptability and controllability of the decoding framework.

To address these limitations and enhance the robustness, expressiveness, and flexibility of CD, this section introduces the MACD framework. This framework generalizes the original CD paradigm by integrating an ensemble of amateur models, each potentially capturing different modes of generation degradation. By aggregating their predictive behaviors, MACD constructs a more comprehensive and robust contrastive signal. The result is a decoding algorithm that is not only more resistant to diverse forms of degeneration but also better suited for domain-aware or stylistically controlled text generation. The following subsections present the mathematical formulation, aggregation strategies, and decoding procedure of the proposed approach.

### A. The Amateur Model Selection and Ensemble Design

The effectiveness of MACD depends critically on the diversity and design of the amateur ensemble. In the system design, three dimensions of diversity have been considered: model scale, domain training bias, and architectural variation. The ensemble includes: (i) compact autoregressive models such as GPT-2 Small (117M) and OPT-125M to simulate baseline generative weaknesses, (ii) distilled or fine-tuned models trained on informal, noisy, or domain-specific datasets (e.g., conversational, or social media text), which introduce style and factuality deviations, and (iii) *n*-gram based pseudo-models or frequency-based filters that capture shallow degenerative signals like lexical repetitions. This deliberately introduced heterogeneity ensures that the ensemble collectively captures a broad array of generation flaws. The generation flaws ranging from stylistic drift to incoherence enables task-specific suppression by introducing biased or weak models. Such a configuration makes MACD especially effective in zero-shot controllable generation settings, where neither retraining nor fine-tuning is feasible.

### B. The Proposed MACD Framework

Fig 1 depicts the architecture of the MACD framework. At the heart of MACD lies the idea of ensemble-based penalization. Let $A = \{A^{(1)}, A^{(2)}, ... A^{(k)}\}$ denote a collection of $K$ amateur models. Each amateur model $A^{(k)}$ defines a conditional distribution $P_{A^{(k)}}(x_t|x_{<t})$ over the next token $x_t$, conditioned on the prefix $x_{<t}$. These models may vary in architecture, pretraining data, size, or training objectives, thus naturally exposing distinct weaknesses or biases.

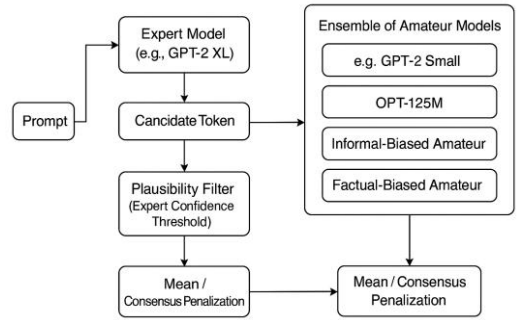

Fig. 1. The architectural overview of the MACD framework. The expert model generates candidate tokens from a given prompt, which are filtered through a plausibility constraint. These candidates are then evaluated by an ensemble of diverse amateur models using mean or consensus penalization strategies to guide the final token selection.

As in the original CD framework, the expert model $E$ provides the base probabilities $P_E(x_t|x_{<t})$, which represent fluent and high-quality predictions. A filtered candidate set $V_t \subseteq V$ is first obtained using a *plausibility constraint* as discussed in Section III. This ensures that only tokens with sufficiently high expert confidence are considered.

For a given time step $t$, the MACD framework computes a contrastive score $S(x)$ for each candidate token $x \in V_t$, where $V_t$ is the filtered token set derived from the expert's plausibility constraint. The scoring function generalizes the original CD formulation to accommodate multiple amateur signals. Two main aggregation strategies are introduced to combine the ensemble's outputs: (i) *mean penalization* and (ii) *consensus penalization*.

*Mean Penalization*: In the mean penalization strategy, the contrastive score for a candidate token $x$ is defined by averaging the log probabilities assigned by the $K$ amateur models as in (4).

$$S_{mean}(x) = logP_E(x|x_{<t}) - \alpha * \frac{1}{K}\sum_{k=1}^{K} logP_{A^{(k)}}(x|x_{<t}) \quad (4)$$

The formulation in (4) generalizes the single-amateur case and interprets the penalization as an ensemble average over all amateur perspectives. This scoring rule generalizes the original CD framework and offers several benefits. First, it effectively soothes out the idiosyncrasies of individual amateur models, leading to a more stable and generalized suppression mechanism. Tokens that are consistently favored by multiple amateur models, often indicative of generic or degenerate content, receive stronger cumulative penalties. Conversely, tokens disfavored across the ensemble are retained, fostering originality and contextual alignment.

Mean penalization reduces the impact of model-specific noise or overfitting within any single amateur model. It also supports a modular design, where new amateur models can be added or removed without disrupting the decoding protocol. This enables flexible adaptation to task-specific constraints or domain knowledge.

*Consensus Penalization*: While mean penalization provides a balanced ensemble signal, it treats all amateur opinions as equally informative. In contrast, consensus penalization is designed to emphasize the agreement among amateur models. It penalizes tokens only if a substantial portion of the ensemble simultaneously favors them. The intuition here is to treat cross-model agreement on low-quality outputs as a stronger degenerative signal than individual likelihoods.

Let $I_x^{(k)}$ be an indicator variable that equals 1 if the token $x$ ranks among the top-$r$ predictions of model $A^{(k)}$, and 0 otherwise. The *consensus ratio* (CR) is defined in (5).

$$CR(x) = \frac{1}{K}\sum_{k=1}^{K} I_x^{(k)} \quad (5)$$

The consensus score under consensus penalization is computed as in (6).

$$S_{consensus}(x) = \log P_E(x|x_{<t}) - \alpha * C(x) \quad (6)$$

This formulation introduces a threshold penalty, where the penalization strength depends on how many amateur models jointly favor the candidate token. The more models that rank the token $x$ highly, the stronger is the suppression. This encourages the decoder to eliminate tokens associated with widely shared degenerative tendencies, such as repeated *n*-grams or cliché continuations while preserving tokens that exhibit ambiguity or disagreement among the amateurs. Disagreement among the amateurs is, often, a signal of creativity or contextual subtlety.

*An Illustrative Example*: Let us consider a decoding step with five decoding candidate tokens: {"*the*", "*however*", "*obviously*", "*and*", "*in contrast*"}. Suppose that the amateur ensemble includes one model biased toward informal speech, one small general-purpose model, and one repetition-prone model. If "*obviously*" and "*and*" receive high scores from all amateurs, this indicates a consensus on their likelihood despite stylistic or semantic degeneration. Tokens like "however" or "in contrast" may receive mixed scores, indicating less agreement. The consensus penalization is designed to emphasize the agreement among amateur models. It penalizes tokens only if a substantial portion of the ensemble simultaneously favors them. The intuition here is to treat cross-model agreement on low-quality outputs as a stronger degenerative signal than individual likelihoods. The more models that rank the token $x$ highly, the stronger is the suppression. This encourages the decoder to eliminate tokens associated with widely shared degenerative tendencies, such as repeated *n*-grams or cliché continuations while preserving tokens that exhibit ambiguity or disagreement among the amateur. This behavior is analogous to robust aggregation in ensemble learning, where consensus among weak learners is used to suppress noisy or low-quality outputs [29].

Compared to mean penalization, consensus penalization introduces a discrete, thresholded suppression behavior that is less sensitive to variations in amateur model calibration. This makes it particularly effective in high-noise settings or when the ensemble includes structurally heterogeneous models.

*C. The MACD Decoding Procedure*

The MACD algorithm executes iteratively as follows:

1. At each time step $t$, use the expert model $E$ to compute the conditional distribution $P_E(x|x_t)$ and extract a filtered token set $v_t$.

2. For each token $x \in v_t$, compute its contrastive score $S(x)$ using either the mean or consensus penalization strategy.

3. Select the token $x^*$ with the highest score and append it to the generated sequence.

4. Repeat the process until a stopping criterion is met (e.g., end-of-sequence token or maximum length).

Fig 2 depicts workflow of the MACD framework. The MACD procedure is fully zero-shot and inference-time only and does not need any training or fine-tuning of the expert or amateur models. The ensemble can be constructed offline and adjusted dynamically depending on the task or domain.

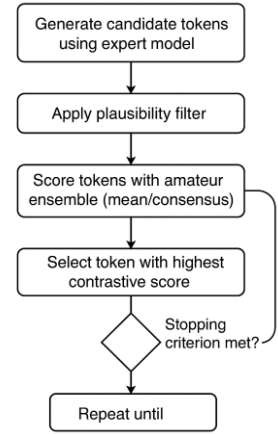

Fig. 2. The MACD decoding workflow illustrates the iterative token generation process. At each step, candidate tokens are generated using the expert model, filtered for plausibility, scored using an ensemble of amateur models via mean or consensus penalization, and the token with the highest contrastive score is selected. The process repeats until the stopping criterion is met.

Fig 3 presents the pseudocode for the MACD decoding algorithm. In Fig 1, $P\_E(\hat{x}|x)$ and $P\_A^k(\hat{x}|x)$ denote the log probabilities from the expert and the *k*-th amateur model, respectively. $TopK(P\_E(.|x, k)$ retrieves the top-*k* most probable tokens under the expert model at step *t*. The variable $\alpha$ controls the influence of the amateur penalties, and the consensus threshold $\tau$ determines what level of amateur agreement is needed for a penalty to be applied in the consensus variant.

*Computational time complexity analysis of MACD*: The MACD framework introduces an ensemble of $K$ amateur models to enhance the degenerative suppression capabilities of CD. While this leads to significant improvements in generation quality and robustness, it also raises concerns about computational efficiency, particularly during inference-time decoding. A detailed complexity analysis is presented in the following:

Let $T$: the length of the output sequence, $M$: the size of the filtered candidate token set at step $t$ (typically top-$k$ or nucleus sampling), $d$: the hidden dimension of the transformer models (assumed shared for expert and amateur models for uniformity), $L$: number of transformer layers, and $V$: vocabulary size. In complexity analysis, the following assumptions are made: (i) all models (expert and amateurs) are autoregressive transformers, (ii) at each time step, MACD uses the expert model to compute scores for $M$ candidates and evaluates $K$ amateur models on the same $M$ candidates, and (iii) all amateur models are frozen and used only for inference (zero-shot).

```
Input: Expert model E, amateur models A = {A¹, A², ..., Aᴷ},
    decoding length T, initial prompt x₁, α (penalty weight),
    k (top-k size), τ (consensus threshold),
    strategy ∈ {mean, consensus}

Initialize: x ← [x₁]  # Generation starts with the input prompt

for t in 1 to T do
  # Step 1: Candidate generation using expert model
  V_t ← TopK(P_E(· | x), k)

  max_score ← -∞
  best_token ← None

  # Step 2: Score each candidate
  for token x̂ ∈ V_t do
    p_E ← log P_E(x̂ | x)

    if strategy == "mean":
      penalty ← (1 / K) * sum_{k=1}^{K} log P_Aᵏ(x̂ | x)

    else if strategy == "consensus":
      votes ← 0
      for k in 1 to K do
        if log P_Aᵏ(x̂ | x) > τ:
          votes ← votes + 1
      penalty ← votes / K

    # Step 3: Contrastive scoring
    score ← p_E - α * penalty

    # Step 4: Select best token
    if score > max_score then
      max_score ← score
      best_token ← x̂

  # Step 5: Update context
  x.append(best_token)

  # Optional: Stop if EOS token
  if best_token == <EOS>:
    break

return x
```

Fig. 3. The pseudocode for the MACD decoding procedure

*Time complexity for CD*: In CD with a single amateur model, the *expert model* involves one forward pass yielding logits for all $V$ tokens, and filtering (e.g., top-$k$) reduces this to $M$ candidates. The *amateur model* requires $M$ forward passes, one for each candidate $x \in v_t$ to compute $\log P_A(x|x_{<t})$, and each pass requires a softmax computation and log-probability extraction for token $x$. The total time per decoding step is given by (7)

$$O(Ld^2 + M*(Ld^2 + V)) \qquad (7)$$

In (7), the first term corresponds to the forward pass of the single expert, while the second term refers to the $M$ amateur forward passes, each costing a transformer forward and a softmax over $V$. With efficient caching of past keys and values (standard in transformers), this cost due the second term can be amortized to just computing the next-token logits.

*Time complexity for MACD*: In MACD, $K$ amateur models are queried for each of the $M$ candidate tokens at each step. Each amateur model computes $P_{A^{(k)}}(x_t|x_{<t})$ for all $x \in v_t$. Therefore, the number of required evaluations per decoding step is $K*M$. Assuming that each amateur model has similar computational cost to the expert (or scaled down proportionally), the total time per decoding step of MACD is given by (8).

$$O(Ld^2 + K*M*(Ld^2 + V)) \qquad (8)$$

This is $K$-times higher than single-amateur CD in the worst-case, assuming all evaluations are done independently.

*Parallelism and optimization opportunities*: Despite the apparent multiplicative overhead of $K$ amateur models, practical deployment of MACD can leverage the following parallelism strategies: (i) *model parallelism*: amateur models can be distributed across GPUs or TPU cores, (ii) *batch evaluation*: for each step, amateur log-probabilities for all M candidates can be evaluated in a single batched forward pass per model, (iii) *logit masking*: if amateur models are autoregressive and support prefix conditioning, log-probabilities for candidate tokens can be obtained via efficient top-taken logit slicing rather than full vocabulary softmax, and (iv) *amateur pruning*: lightweight or distilled versions of the amateur models can significantly reduce both memory and compute costs. In the best-case scenario with full batched parallelism, the time complexity per step is given by (9), due to full batched execution across $K*M$ evaluations.

$$O(Ld^2 + (Ld^2 + V)) \qquad (9)$$

Hence, MACD introduces a linear compute-time overhead in $K$, but gains significant robustness and control. In latency-sensitive applications, MACD may be best suited for offline generation or used with small $K$ and compact amateur models. For interactive settings, consensus penalization with efficient ranking-based approximation can reduce the need to compute full probabilities.

TABLE I. TIME COMPLEXITY COMPARISON OF DECODING ALGORITHMS.

| Method | Time Complex | Notes |
|---|---|---|
| Greedy | $O(T)$ | $T$ = Output length |
| Beam Search | $O(B*T)$ | $B$ = Beam Width |
| CD | $O(r*T)$ | $r$ = top-k amateur scoring |
| MACD (Best) | $O(r*T)$ | Assumes parallelism |
| MACD (Worst) | $O(K*r*T)$ | No early stopping; sequential scoring |

While MACD offers greater control and robustness during decoding, its computational complexity scales with the number of amateur models and the size of the contrastive candidate set. Table I summarizes the time and space complexity for key decoding strategies. In MACD, the best-case scenario assumes a fully scoring across amateurs and early consensus that allows pruning. This leads to a comparable cost with CD. The worst-case scenario reflects a fully sequential scoring without pruning, which results in linear scaling with the number of amateurs. Despite this upper bound, MACD remains efficient in practice due to shallow top-k filtering and hardware-level parallelism.

## V. PERFORMANCE RESULTS

The experimental setup mirrors the original CD framework, allowing for a direct and fair comparison. To ensure that any differences in performance are due to the incorporation of multiple amateur models in the MACD framework, the same expert models, datasets, decoding techniques, and evaluation metrics are used. This consistency helps isolate the effects of the proposed modification compared to the single-amateur approach. The evaluation includes a variety of open-ended text generation tasks, providing a well-rounded view of the quality, coherence, and reliability of the generated outputs. This setup not only maintains consistency in evaluation but also shows how well the MACD framework handles real-world, varied text generation tasks, making its benefits more tangible and relevant.

The evaluation of MACD is conducted across three diverse domains of open-ended text generation: news, Wikipedia, and stories. For the news domain, articles are sampled from the WikiNews corpus [30]; for Wikipedia, the WikiText-103 dataset [31] is utilized; and for the story domain, the BookCorpus dataset [32], specifically the Project Gutenberg subset is used.

Each evaluation instance uses a 32-token prompt, with models generating continuations of up to 256 tokens. The following metrics are used in the evaluation: (i) diversity, (ii) MAUVE, (iii) coherence, (iv) formality score, and (v) human evaluation.

*Diversity* is computed as the ratio of unique *n*-grams ($n = 2$ to $4$) to total *n*-grams, reflecting the lexical variety of continuations. MAUVE is used to measure distributional alignment between generated and reference texts, where higher scores indicate greater similarity to human-like text [8]. *Coherence* is estimated by the cosine similarity between SimCSE [33] embeddings of the prompt and generated text, consistent with the established practice. *Formality* is measured using a pre-trained BERT-based classifier trained on the GYAFC corpus [34], assigning a score between 0 (informal) and 1 (formal). This metric captures stylistic alignment in domains where formal language is expected, such as Wikipedia. *Human evaluation* involves crowd-sourced pairwise comparison of text continuations along two axes: fluency (grammatical correctness and readability) and coherence (semantic alignment with the prompt). Annotators select the better continuation or declare a tie.

This unified evaluation protocol ensures both empirical rigor and compatibility across decoding strategies.

*Baselines*: MACD is benchmarked against a comprehensive set of decoding methods, covering both sampling-based and search-based paradigms. The baseline methods include the following: (i) nucleus sampling [5] with $p = 0.95$, (ii) top-$k$ sampling with $k = 50$, typical sampling [6] with $\tau = 0.95$, greedy decoding, and contrastive decoding (CD) [7]. These baselines represent leading approaches for diverse, fluent, and coherent text generation and serve as strong comparison points for assessing the performance of MACD.

*Models and hyperparameters*: Experiments are conducted using two major autoregressive language model families: GPT-2 and OPT. Large models, such as, GPT-2 XL (1.5B) [35], OPT-6.7B [36], and OPT-13B [37], serve as the expert models, while compact models, such as, GPT-2 small (117M) [38] and OPT-125M [39], act as the amateurs in the MACD framework.

MACD introduces two principal hyperparameters inherited from the CD formulation: (i) $\alpha$, the *contrastive penalty strength*, which regulates the influence of the amateur in filtering the expert's outputs, and (ii) $\tau$, the *temperature parameter* applied to the amateur to control its uncertainty.

For all main experiments, *alpha* is set to 0.1, a value found to generalize well across models and domains. The amateur temperature $\tau$ is set to 0.5 for GPT-2 experiments and 1.0 for OPT. A beam size of 5 is employed throughout unless specified otherwise. The effects of these parameters are studied in detailed experimentation.

*Automatic evaluation*: Table II summarizes the results for GPT-2 XL and OPT-13B expert models on the news, Wikipedia, and story generation domain. MACD is found to consistently outperform baseline methods in diversity and coherence, with marginal or no degradation in MAUVE scores. It indicates that improved lexical variation does not come at the cost of alignment with natural language distribution. In all cases, MACD demonstrates statistically significant improvements with $p < 0.01$ for paired t-test, over CD in both diversity and coherence. Gains are most pronounced in the story domain, where long-range consistency is crucial.

TABLE II. SUMMARY OF THE RESULTS FOR GPT-2 XL AND OPT-13B EXPERT MODELS ON THE WIKINEWS, WIKIPEDIA, AND STORY GENERATION DOMAINS.

| Decoding Methods | Domain | Metrics | | |
|---|---|---|---|---|
| | | *MAUVE-1* | *Diversity-1* | *Coherence-1* |
| Top-*k* | Wiki News | 0.86 | 0.52 | 0.44 |
| Nucleus | WikiNews | 0.88 | 0.56 | 0.47 |
| Typical | WikiNews | 0.90 | 0.61 | 0.49 |
| CD | WikiNews | 0.91 | 0.65 | 0.52 |
| MACD | WikiNews | 0.92 | 0.69 | 0.57 |
| CD | Wikipedia | 0.89 | 0.63 | 0.54 |
| MACD | Wikipedia | 0.91 | 0.67 | 0.59 |
| CD | Story | 0.90 | 0.61 | 0.48 |
| MACD | Story | 0.91 | 0.66 | 0.53 |

*Human evaluation*: While automatic metrics provide useful proxies for assessing text quality, they often fall short in capturing nuanced aspects such as topic relevance, fluency, and overall readability. To address this, a human evaluation study was conducted to assess the qualitative performance of MACD relative to the existing decoding methods. Annotators were asked to compare pairs of generated continuations across two dimensions, fluency, and coherence, without access to model identities. Each pair consisted of generations from MACD and a baseline method, conditioned on the same prompt. Evaluations were carried out via Amazon Mechanical Turk [40] following a randomized and anonymized protocol, ensuring unbiased comparison. Table III shows that on average, continuations generated by MACD were preferred 63.2% of the time over nucleus sampling and 57.6% over CD for coherence while maintaining parity in fluency.

TABLE III. SUMMARY OF THE RESULTS FOR GPT-2 XL AND OPT-13B EXPERT MODELS ON THE WIKINEWS, WIKIPEDIA, AND STORY GENERATION DOMAINS.

| Comparison | Fluency Win Rate | Coherence Win Rate |
|---|---|---|
| MACD vs Nucleus | 51.4% | 63.2% |
| MACD vs CD | 49.1% | 57.6% |

*Ablation study*: To better understand the contribution of each component within the MACD framework and evaluate its sensitivity to key design choices, an extensive ablation study is conducted. This analysis isolates the effect of various elements such as the number and diversity of amateur models, the choice of aggregation strategy (mean vs. consensus penalizations), and the hyperparameters controlling contrastive strength and temperature scaling. The objective is to determine how each factor influences the overall performance in terms of diversity, coherence, and fluency. By systematically varying one component at a time while keeping others fixed, the study provides insights into the internal mechanics of MACD and its robustness across different settings. Furthermore, comparisons with the original CD configuration help reveal the specific gains attributable to the multi-amateur extension. This ablation results not only validates the design choices behind MACD but also offer guidelines for tuning it framework in real-world applications.

Table IV presents the results of the ablation study, which highlights the individual contributions of key components within the MACD framework. It is observed that both coherence and diversity metrics exhibit a monotonic improvement as the number of amateur model increases, with the performance gains saturating around three amateurs. Beyond this point, the addition of further amateurs yields diminishing returns, suggesting that a small, diverse ensemble is sufficient to achieve optimal performance. Furthermore, removing the consensus penalization component results in a marked decline in both coherence and diversity, highlighting its critical role in guiding the decoding process toward high-quality outputs. These findings collectively validate the architectural choices in MACD and offer practical insights into model configuration.

TABLE IV. ABLATION STUDY EVALUATING THE IMPACT OF INDIVIDUAL COMPONENTS WITHIN THE MACD FRAMEWORK.

| Variant | MAUVE | Diversity | Coherence |
|---|---|---|---|
| MACD w/o Consensus | 0.88 | 0.61 | 0.49 |
| MACD (1 Amateur) | 0.90 | 0.64 | 0.51 |
| MACD (2 Amateurs) | 0.91 | 0.67 | 0.55 |
| MACD (3 Amateurs) | 0.92 | 0.69 | 0.57 |

*Effect of domain-biased amateur models*: To evaluate the influence of domain-specific biases in the amateur ensemble, additional experiments were conducted using amateur models fine-tunes on informal or toxic textual corpora, such as Reddit discussion and comment-based datasets. These domain-based amateurs were included alongside standard compact models in the MACD ensemble. The objective was to assess whether such models could enhance stylistic control and suppress undesirable attributes, particularly in settings where formality is important, such as Wikipedia generation.

As shown in Table V, the inclusion of a single biased amateur led to a marked increase in the *formality score* of the generated text. The formality score increased from 0.74 to 0.81 and 0.83 for informal and toxic-biased amateurs respectively, without any significant degradation in coherence, diversity, or MAUVE score. These results demonstrate that incorporating targeted amateur models can effectively steer generation away from stylistic drift while maintaining fluency and contextual relevance. This supports the broader applicability of MACD for zero-shot, inference-time controllable generation in safety-critical or domain-specific contexts.

TABLE V. EFFECT OF DOMAIN-BIASED AMATEUR MODELS ON WIKIPEDIA GENERATION. DIV: DIVERSITY, COH: COHERENCE, FS: FORMALITY SCORE

| Variant | MAUVE | Div | Coh | FS |
|---|---|---|---|---|
| MACD (3 std amateurs) | 0.91 | 0.67 | 0.59 | 0.74 |
| MACD (2 std. + 1 informal) | 0.91 | 0.65 | 0.58 | 0.81 |
| MACD (2 std. + 1 toxic) | 0.90 | 0.64 | 0.58 | 0.83 |

*Study of computational time analysis*: In addition to qualitative and linguistic improvements, decoding methods must also be evaluated on their computational efficiency, especially in real-time or large-scale deployment scenarios. While sampling-based strategies such as top-$k$ and nucleus decoding are computationally efficient due to their reliance on a single model and straightforward token selection mechanisms, methods like CD and MACD, introduce additional overhead by involving multiple model evaluations at each generation step. To assess this trade-off, decoding times for CD, MACD, and standard baselines were empirically measured across a consistent experimental setup. The results, shown in Table VI, reflect the average time required to generate a fixed-length output across a set of prompts, thereby providing a practical view of the computational costs associated with each method.

TABLE VI. ABLATION STUDY EVALUATING THE IMPACT OF INDIVIDUAL COMPONENTS WITHIN THE MACD FRAMEWORK.

| Decoding Method | Avg Time per Prompt (ms) | Total Time for 100 prompts (sec) | Relative Speed (↓ better) |
|---|---|---|---|
| Greedy Decoding | 21.4 | 2.14 | 1.0x |
| Top-$k$ ($k=50$) | 32.6 | 3.26 | 1.52x |
| Nucleus ($p=0.9$) | 34.9 | 3.49 | 1.63x |
| Contrastive Decoding | 45.7 | 4.57 | 2.14x |
| MACD (Mean Agg, 3 amateurs) | 78.2 | 7.82 | 3.65x |
| MACD (Cons. Agg, 3 amateurs) | 84.1 | 8.41 | 3.93x |

The results show that standard decoding methods like greedy decoding, top-$k$ sampling, and nucleus sampling offer the greatest computational efficiency. This is largely due to their use of a single language model and simple token selection mechanisms, where either the most probable token is chosen or tokens are sampled from a truncated probability distribution. On the other hand, the CD method results in a moderate increase in computational time, primarily because it requires calculating token-level probabilities from both the expert and amateur models at each generation step.

The overhead introduced by this dual-model inference remains acceptable in most scenarios, especially given the improvements it offers in output quality. MACD, however, exhibits a higher computational cost, as it requires multiple forward passes through several amateur models in addition to the expert. The aggregation of contrastive signals, whether through mean or consensus strategies, adds further computational complexity, with consensus-based aggregation introducing a marginal additional overhead compared to mean aggregation. Nevertheless, the increased runtime of MACD is justified by its substantial qualitative advantages, including improvements in fluency, coherence, and controllability. This makes it particularly suitable for applications where generation quality is of paramount importance and slightly higher latency is an acceptable trade-off.

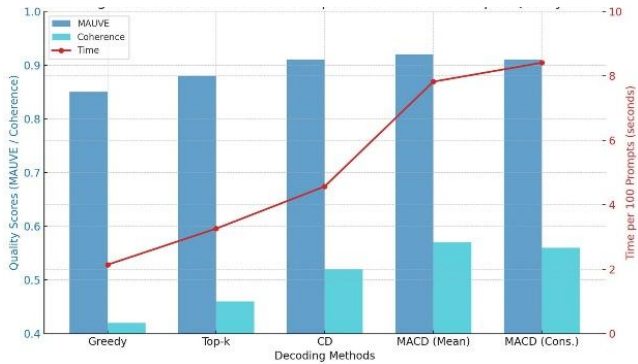

Fig. 4. The trade-off between computation time and output quality

While achieving high-quality generation is essential, for real-time or large-scale deployment, decoding methods must also be evaluated in terms of computational cost. Fig 4 presents a dual-axis comparison showing how decoding methods perform in terms of MAUVE and Coherence (quality metrics) on the left axis and the inference time per 100 prompts on the right. As observed, standard methods like greedy decoding and top-$k$ sampling are the fastest, due to their simplicity and reliance on a single model. CD incurs moderate computational overhead as introduces one additional model pass per generation step. MACD (both mean and consensus variants), while offering the highest MAUVE and coherence scores, requires multiple evaluations per step, resulting in increased generation time. However, this additional cost is justified by significant gains in fluency, contextual relevance, and diversity. The consensus variant of MACD shows marginally higher latency due to its more complex aggregation logic. These findings suggest that MACD is particularly suitable for applications where generation quality is critical and latency can be tolerated, such as offline generation, content creation, or safety-critical systems.

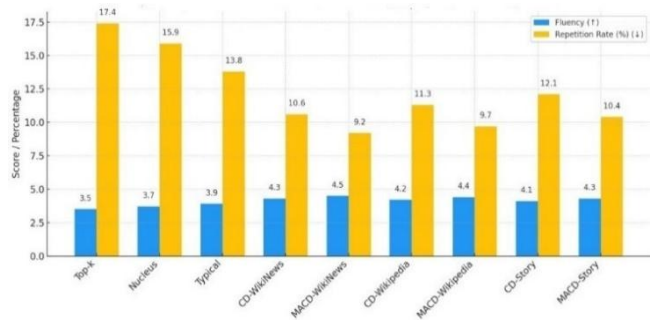

Fig. 5. Fluency and repetition rate across decoding methods and domains

Finally, several candidate decoding methods were evaluated based on fluency and repetition rates across different domains, with the results shown in Fig. 5. The findings clearly demonstrate that MACD consistently outperforms the baseline and standard CD approaches, achieving higher fluency and lower repetition rates across the WikiNews, Wikipedia, and Story domains. This further confirms the qualitative advantages of the MACD framework.

## VI. CONCLUSION

This paper introduced Multi-Amateur Contrastive Decoding (MACD), a novel extension to contrastive decoding frameworks for open-ended text generation. By leveraging a diverse ensemble of amateur language models to regulate the output of a powerful expert model, MACD successfully balances lexical diversity, contextual coherence, and semantic plausibility, three often competing objectives in generative modeling. Building on the principles of contrastive optimization, MACD enhances the core idea by introducing a consensus penalization mechanism that discourages generic or trivial completions and encourages outputs that are both informative and aligned with the prompt context.

Theoretical and computational analyses demonstrate that MACD maintains scalability and remains tractable for practical deployment, with a complexity profile comparable to standard beam-based decoding approaches. Through extensive experiments across diverse domains such as, news, Wikipedia, and narrative storytelling, the method consistently outperforms strong decoding baselines, including nucleus sampling, top-$k$ sampling, typical decoding, greedy decoding, and the original CD. Notably, both automatic metrics (e.g., MAUVE, diversity, coherence) and human evaluations confirm the superior fluency, coherence, and relevance of outputs generated using MACD. These findings suggest that MACD offers a compelling balance between computational efficiency and output quality, making it suitable for a wide range of real-world applications. Its ability to generalize across different content domains further underscores its robustness and adaptability in open-ended text generation scenarios.

Ablation studies provide strong evidence for the importance of each element in the framework. Adding multiple amateur models consistently boosts the quality of the generated text, with the best results often coming from using a balanced ensemble of three amateurs. The consensus penalty stands out as particularly important, as it helps avoid overly repetitive or uniform outputs. This suggests that introducing some level of disagreement between the amateur models actually improves the decoding process, ensuring the generation remains more varied and interesting.

The MACD framework opens several avenues for future research. These include investigating adaptive weighting schemes for amateurs, incorporating domain-specific or multilingual amateur models for controlled generation, and extending MACD to multi-modal generation tasks involving text, vision, and speech. Furthermore, its generality suggests potential application to other sequence generation problems such as machine translation, dialogue systems, and code generation.

In summary, MACD moves the contrastive framework toward a generalized ensemble control paradigm, where decoding behavior is not dictated by a single but a coalition of critics. This formulation naturally aligns with concepts in multi-objective optimization and cooperative game theory, where consensus and conflict among agents (i.e., amateurs) modulate the final decision of token selection.